\documentclass{article} %
\usepackage{colm2024_conference}

\usepackage{booktabs}
\usepackage{graphicx}
\usepackage{enumitem}
\usepackage{wrapfig}
\usepackage{algorithm}
\usepackage{algpseudocode}

\usepackage{microtype}
\usepackage{amsmath}
\usepackage{colortbl}
\usepackage[utf8]{inputenc}
\definecolor{lightgray}{rgb}{0.9,0.9,0.9}
\usepackage{caption}
\usepackage{subcaption}
\usepackage{xcolor}
\usepackage{setspace}
\usepackage{url}
\usepackage{multirow}
\usepackage{colortbl}
\usepackage{tabularx}
\usepackage{blindtext}
\usepackage{pgfplots}
\pgfplotsset{compat=1.18} 
\usepackage{tikz}
\usetikzlibrary{er,positioning,bayesnet}
\usepackage{makecell}
\usepackage{tipa}
\usepackage{siunitx}
\usepackage{nicefrac}
\usepackage{tocloft}
\usepackage{listings}
\usepackage[raster,skins]{tcolorbox} %
\usepackage{xltabular}
\usepackage{adjustbox}
\usepackage{xurl}
\usepackage{amsmath,amsfonts,bm}

\def\eqref#1{equation~\ref{#1}}
\def\1{\bm{1}}

\DeclareMathAlphabet{\mathsfit}{\encodingdefault}{\sfdefault}{m}{sl}
\SetMathAlphabet{\mathsfit}{bold}{\encodingdefault}{\sfdefault}{bx}{n}

\title{Ovis-Image Technical Report}

\author{
\bf Ovis Team, Alibaba Group
}

\begin{document}

\maketitle

\begin{abstract}
We introduce \textbf{Ovis-Image}, a 7B text-to-image model specifically optimized for high-quality text rendering, designed to operate efficiently under stringent computational constraints. Built upon our previous Ovis-U1 framework, Ovis-Image integrates a diffusion-based visual decoder with the stronger Ovis 2.5 multimodal backbone, leveraging a text-centric training pipeline that combines large-scale pre-training with carefully tailored post-training refinements. Despite its compact architecture, Ovis-Image achieves text rendering performance on par with significantly larger open models such as Qwen-Image and approaches closed-source systems like Seedream and GPT4o. Crucially, the model remains deployable on a single high-end GPU with moderate memory, narrowing the gap between frontier-level text rendering and practical deployment. Our results indicate that combining a strong multimodal backbone with a carefully designed, text-focused training recipe is sufficient to achieve reliable bilingual text rendering without resorting to oversized or proprietary models.
\end{abstract}

\vspace{-15pt}
\begin{figure}[hbp]
    \centering
    \includegraphics[width=\textwidth]{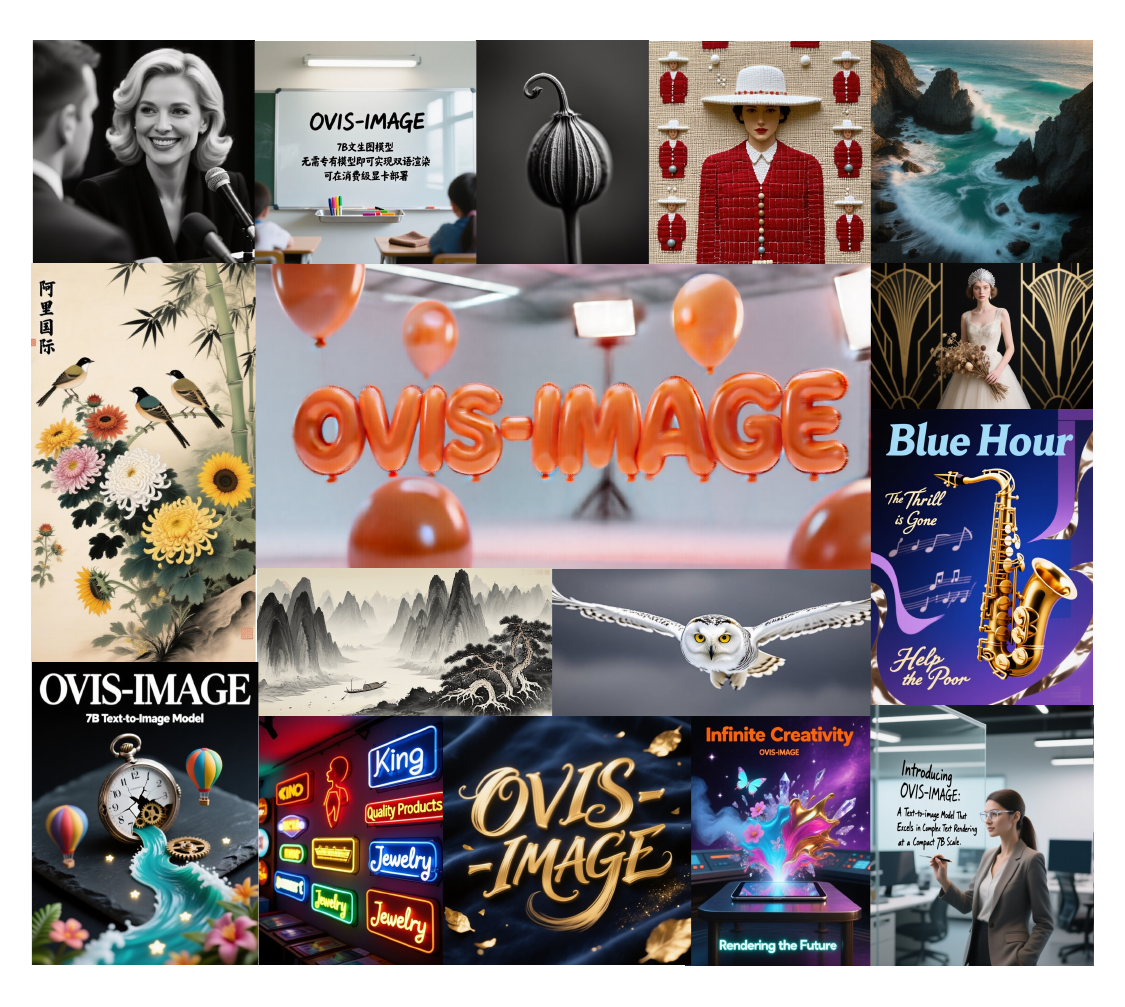}
    \caption{Comprehensive illustration of the functional capabilities of Ovis-Image.}
    \label{fig:intro}
\end{figure}

\newpage

\section{Introduction}
\label{sec:intro}

Recent breakthroughs in image generation have significantly enhanced both the visual fidelity and controllability of synthesized images, exemplified by powerful text-to-image models~\citep{qwen_image,HunyuanImage-2.1} and unified multimodal frameworks~\citep{chatgpt4o,ovis_u1,nano_banana_pro}. Nevertheless, despite this rapid progress, achieving reliable, high-quality text rendering within images at low computational cost remains a persistent challenge. It requires models to simultaneously master fine-grained visual synthesis and robust language comprehension. In practice, strong text-rendering capabilities are typically found only in very large models~\citep{qwen_image,HunyuanImage-2.1}, which are difficult to deploy, or in closed-source systems~\citep{seedream4,nano_banana_pro,chatgpt4o}, which hinder integration, customization, and reproducibility.

In our prior work, \citet{ovis_u1} introduced \emph{Ovis-U1}, a 3B unified model that integrates multimodal understanding, text-to-image generation, and image editing within a single framework. It achieves competitive performance on multimodal understanding and generation benchmarks, approaching the behavior of proprietary models like GPT4o~\citep{chatgpt4o} in many practical scenarios. Nonetheless, with limited parameters, Ovis-U1 struggles with artifacts and hallucinations, and its in-image text quality still lags behind the best closed-source generators~\citep{seedream4,nano_banana_pro}.

Concurrently, recent text-to-image systems~\citep{qwen_image,HunyuanImage-2.1} with especially strong text rendering ability tend to follow two patterns. First, they are tightly integrated with powerful multimodal understanding backbones~\citep{bai2025qwen2.5vl}, leveraging improved visual perception and OCR-like capabilities to reason about embedded text. Second, they scale model size to tens of billions of parameters. This combination yields impressive text-centric performance but substantially increases the cost of training and deployment in real-world applications. Motivated by these observations, we aim to design a specialized generator that prioritizes text rendering while preserving solid visual fidelity on general concepts and keeping computational cost acceptable.

Within this design space, we present \textbf{Ovis-Image}, a 7B text-to-image model that narrows the gap between efficiency and capability. We retain the successful pattern of coupling a strong multimodal backbone with a diffusion-based visual decoder, upgrade the multimodal backbone to the more capable Ovis 2.5~\citep{ovis25}, and scale the vision-side generator to 7B parameters. Despite its compact size, Ovis-Image delivers text rendering performance comparable to much larger 20B-class open models such as Qwen-Image~\citep{qwen_image} and approaches state-of-the-art closed-source generators like GPT4o and Gemini~\citep{nano_banana_pro} on a range of text-centric tasks. In practice, it produces sharp, legible, and semantically consistent text for posters, banners, UI mockups, and scenes, while maintaining competitive image quality and prompt adherence on general-purpose generation.

In summary, Ovis-Image offers the following key advantages:

\begin{itemize}
    \item \textbf{Strong text rendering at a compact 7B scale.}
    Ovis-Image is a 7B text-to-image model that delivers text rendering quality comparable to much larger 20B-class systems such as Qwen-Image and competitive with leading closed-source models like GPT4o in text-centric scenarios, while remaining small enough to run on widely accessible hardware.

    \item \textbf{High fidelity on text-heavy, layout-sensitive prompts.}
    The model excels on prompts that demand tight alignment between linguistic content and rendered typography (e.g., posters, banners, logos, UI mockups, infographics), producing legible, correctly spelled, and semantically consistent text across diverse fonts, sizes, and aspect ratios without compromising overall visual quality.

    \item \textbf{Efficiency and deployability.}
    With its 7B parameter budget and streamlined architecture, Ovis-Image fits on a single high-end GPU with moderate memory, supports low-latency interactive use, and scales to batch production serving, bringing near–frontier text rendering to applications where tens-of-billions–parameter models are impractical.
\end{itemize}

Taken together, these advantages indicate that high-quality in-image text rendering does not inherently require extremely large models. With appropriate architectural choices and a text-centric training pipeline built on top of a strong multimodal backbone, it is possible to approach frontier-level text-to-image performance within a compact 7B footprint.

\section{Architecture}

\begin{figure}
  \centering
  \includegraphics[width=\linewidth]{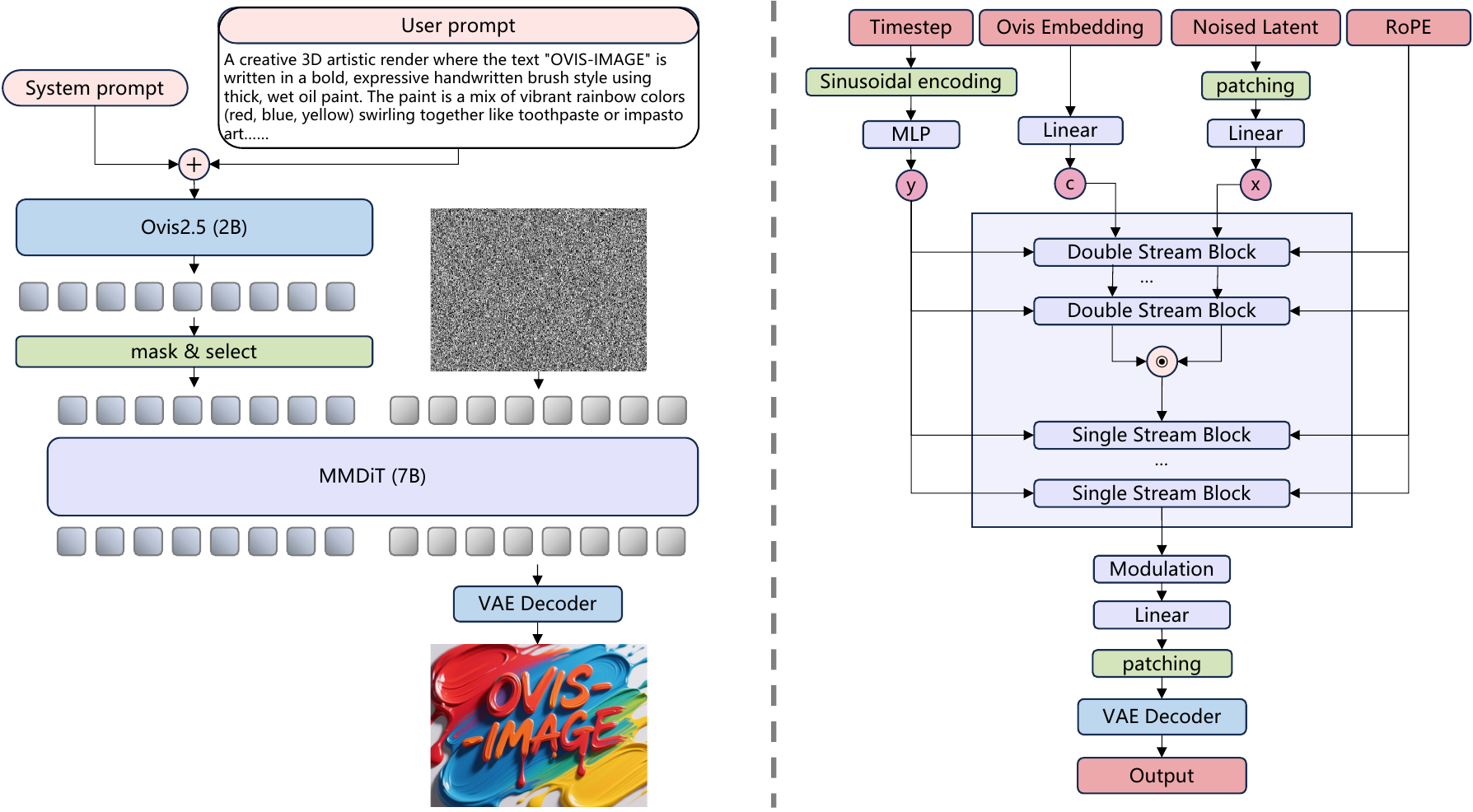}
  \caption{The overall architecture of Ovis-Image. The architecture of Ovis-Image builds upon Ovis-U1, enhancing its capabilities by increasing the parameters of MMDiT and streamlining the structural design to create a more efficient and refined overall framework. }
  \label{fig:framework}
\end{figure}

The architecture of Ovis-Image is presented in Fig.~\ref{fig:framework}. Ovis-Image builds upon the architecture of Ovis-U1~\citep{ovis_u1}, simplifying certain structures while increasing the parameters of the MMDiT backbone. A detailed summary of each module is provided in Tab.~\ref{tab:model_detail}.

\textbf{MMDiT \& VAE.} 
Building upon Ovis-U1, we employ MMDiT \citep{esser2024scaling} with RoPE \citep{su2024roformer} as the visual decoder and use flow matching as the training objective. Inspired by Flux.1 Lite~\citep{flux1-lite}, Ovis-Image incorporates a structure of 6 double-stream blocks and 27 single-stream blocks. To enhance model capacity, the number of attention heads has been increased to 24. SwiGLU~\citep{dauphin2017language} is utilized as the activation function. Additionally, we integrate the VAE model from FLUX.1-schnell~\citep{flux2024} and keep it frozen throughout the training process.

\textbf{Text Encoder.} 
We use the Ovis~\citep{lu2024ovis} series as the text encoder in our framework. Unlike large language models like Qwen, Ovis is specifically trained on multimodal datasets, enabling superior alignment between visual and textual representations. To enhance computational efficiency, we adopt Ovis2.5-2B \citep{ovis25}, which demonstrates better performance than Qwen2.5-VL-7B \citep{bai2025qwen2.5vl} on the OpenCompass benchmark suite. Unlike Ovis-U1, Ovis-Image simplifies the architecture by removing the refiner structure and directly utilizing the final hidden states of Ovis as the conditioning input for image generation.

\begin{table}
  \caption{The model structure details of Ovis-Image.}
  \label{tab:model_detail}
  \centering
  \small
  \begin{tabular}{lrl}
    \toprule
    \textbf{Module} & \textbf{\#Param. (B)} & \textbf{Pretrain} \\
    \midrule
    MMDiT & 7.37 & - \\
    Text Encoder & 2.57 & AIDC-AI/Ovis2.5-2B \\
    VAE & 0.08 & black-forest-labs/FLUX.1-schnell \\
    \midrule
    Total & 10.02 & \\
    \bottomrule
  \end{tabular}
\end{table}

\section{Data Composition}

\subsection{Data for Pre-training}
\label{subsec:pretrain_data}

We pre-train Ovis-Image on a large, heterogeneous corpus of image--text pairs drawn from a mixture of web-scale, licensed, and synthetic sources. The corpus covers everyday photographs, illustrations, design assets, and UI-like mockups, with descriptions ranging from short captions to instruction-style prompts. To better align the text with the visual content, we perform large-scale recaptioning in both Chinese and English. In addition to generic visual content, we include data slices where text is a salient visual element, such as posters, banners, logos, and UI layouts.

To improve data quality, we apply a multi-stage filtering pipeline combining simple heuristics, lightweight models, and cross-modal consistency checks to remove corrupted images, severely mismatched or uninformative captions, and content that does not satisfy basic safety and policy requirements. We further perform coarse deduplication to reduce near-duplicate images and prompts. For text rendering in particular, we augment the corpus with synthetic samples generated by a rendering engine that composes clean typographic text into diverse backgrounds and layouts, providing the model with controlled examples of fonts, sizes, and placements.

\subsection{Data for Supervised Fine-tuning}

For supervised fine-tuning, we curate a higher-quality subset of image-text pairs, emphasizing clean visuals and well-formed prompts. Compared to the pre-training mixture, this corpus shifts toward higher-resolution images (typically at $1024$ pixels) and covers a broad range of aspect ratios to better match real-world use cases. In addition to natural images, we include a moderate amount of synthetic data, which provides sharper details, more controlled layouts, and richer coverage of rare concepts. We perform simple balancing over content type, resolution, and aspect ratio to avoid overfitting to a few dominant patterns. Overall, this stage refines the model’s ability to produce high-fidelity, instruction-following generations under consistent high-resolution constraints.

\subsection{Data for DPO}

For the DPO stage, we construct a preference dataset on top of the supervised distribution. About 90\% of this pool consists of high-quality generations that cover common object categories and everyday scenes with strong aesthetic quality. These images are pre-filtered using an ensemble of automatic scorers (including HPSv3~\cite{ma2025hpsv3}, CLIP~\cite{clip}, PickScore~\cite{pap}, and related metrics), so that only samples with both good visual appeal and reasonable prompt alignment are retained. The remaining $\sim$10\% comes from an in-house collection focusing on design and creative content (e.g., posters, illustrations, and stylized compositions), which exposes the model to more structured layouts and non-photographic styles. For every selected prompt, we take the associated high-quality image from this pool as one candidate and generate a second candidate with the SFT model under the same text conditioning. Both images are then evaluated by multiple scoring models, and their scores are combined into an overall ranking. The higher-scoring image is treated as the winner and the other as the loser, yielding a preference pair for that prompt.

\subsection{Data for GRPO}

The GRPO stage operates on a prompt distribution that is deliberately different from the one used for DPO. Instead of broad, general-purpose generation prompts, we focus on a compact set of text-rendering prompts that stress the model's ability to place and stylize text in images. These prompts cover both Chinese and English, span a range of fonts and layouts (for example, posters, title cards, UI elements, and product labels), and vary in difficulty from short slogans to longer multi-line phrases. By concentrating the budget on this slice of the space, the GRPO data directly targets one of the known weaknesses of diffusion models, namely accurate and legible text generation.

\section{Training}

\subsection{Training Infrastructure}

Our training framework is built with PyTorch, utilizing Hybrid Sharding Data Parallel (HSDP) for efficient data parallelism and parameter sharding. To address memory limitations in larger models, we incorporate gradient checkpointing and activation offloading \citep{korthikanti2023reducing}. Training is conducted with bfloat16 (BF16) mixed precision while maintaining FP32 master weights for accuracy. To optimize training efficiency, we employ Flash Attention \citep{dao2022flashattention,dao2023flashattention2} and regional compilation techniques. Additionally, distributed checkpointing is implemented to minimize the overhead of saving model state.

\subsection{Training Procedure}

Ovis-Image is trained using a four-stage pipeline: a pretraining stage followed by three post-training stages. Each subsequent stage is initialized using the checkpoint from the previous stage.

\textbf{Stage 0: Pretraining.} 
The MMDiT is initialized randomly and optimized throughout all four training stages, while the text encoder and VAE use pretrained weights and remain frozen during training. The training objective follows the standard noise-prediction loss commonly applied in flow-matching-style diffusion models~\citep{esser2024scaling}. AdamW serves as the optimizer, paired with a constant learning rate schedule and a brief linear warmup period. Initially, the model is trained on $256\times 256$ images, followed by training on images of varying resolutions and aspect ratios, ranging from 512 to 1024 pixels and 0.25 to 4.0, respectively. This process produces a robust initial model, which is further refined in the later stages using supervised data and preference optimization.

\textbf{Stage 1: Supervised Fine-tuning.}
In the second stage, we move from generic caption data to instruction-style supervision tailored to common text-to-image usage. Starting from the pretraining checkpoint, we finetune the MMDiT on a mixture of open and proprietary datasets. This stage therefore teaches the model not only what to draw, but also how to interpret instruction-like descriptions, constraints, and text-rendering requirements.

The training objective remains the same noise-prediction loss as in pretraining, applied to latent representations of images up to $1024$ resolution with different aspect ratios, so that the model learns to handle variable input sizes and aspect ratios at inference time. We use a small learning rate and a shorter schedule, which helps preserve the general visual competence learned during pretraining while adapting to the instruction-style and text-rendering distributions. In practice, we find that this supervised tuning substantially improves faithfulness to user prompts and aesthetic quality of the generated images.

\textbf{Stage 2: DPO.}
In the second stage, we apply Direct Preference Optimization (DPO)~\citep{wallace2024diffusion} directly to the diffusion model using a mixture of human and model generated preference data. Each training example consists of a prompt $c$ and two images $(x^{w}, x^{\ell})$, where $x^{w}$ is labeled as preferred (winner) and $x^{\ell}$ as dispreferred (loser). We keep a frozen reference model $p_{\text{ref}}$ at the end of the supervised stage and treat the current image decoder $p_{\theta}$ initialized from $p_{\text{ref}}$ as the policy model, which is trained to assign higher probability to the denoising trajectory leading to the preferred sample.

For each pair, we compute a DPO-style log-likelihood ratio
\begin{equation}
    \Delta \log p_{\theta}(x^{w}, x^{\ell} \mid c)
    = \bigl[\log p_{\theta}(x^{w} \mid c) - \log p_{\theta}(x^{\ell} \mid c)\bigr]
      - \bigl[\log p_{\text{ref}}(x^{w} \mid c) - \log p_{\text{ref}}(x^{\ell} \mid c)\bigr],
\end{equation}
and minimize the standard Diffusion-DPO objective
\begin{equation}
    \mathcal L_{\mathrm{DPO}}(\theta)
    = - \mathbb{E}_{(c, x^{w}, x^{\ell})}
      \Bigl[\log \sigma\bigl(\beta \, \Delta \log p_{\theta}(x^{w}, x^{\ell} \mid c)\bigr)\Bigr],
\end{equation}
where $\sigma(\cdot)$ is the logistic function and $\beta$ is a temperature hyperparameter. In practice, the log probabilities are instantiated via the diffusion reconstruction losses along the denoising trajectories, following~\citet{wallace2024diffusion}.

Following Diffusion-SDPO~\citep{fu2025diffusion}, we further incorporate a winner-preserving safeguard that modifies how the winner and loser branches contribute to the update. Let $\mathcal L^{w}(\theta)$ and $\mathcal L^{\ell}(\theta)$ denote the per-pair diffusion reconstruction losses for the winner and loser, and let $g^{w} = \nabla_{o} \mathcal L^{w}$ and $g^{\ell} = \nabla_{o} \mathcal L^{\ell}$ be their output-space gradients. Diffusion-SDPO explicitly computes a gradient scale factor $\lambda_{\text{safe}}$ to stabilize the optimization as follows:
\begin{equation}
    \lambda_{\text{safe}}
    = \operatorname{clip}\left(
        \mu \,\frac{\langle g^{w}, g^{\ell} \rangle}
                 {\lVert g^{\ell} \rVert^{2} + \varepsilon},
        \lambda_{\min}, \lambda_{\max}
    \right),
\end{equation}
where $\mu$ controls the overall strength of the loser branch, $\varepsilon$ is a small constant for numerical stability, and the clipping interval $[\lambda_{\min}, \lambda_{\max}]$ is chosen so that the first-order change of the winner loss is non-positive. Intuitively, the loser gradient is down-weighted whenever it conflicts with the winner gradient, which implicitly clips overly aggressive loser updates and preserves the quality of the preferred branch. Omitting this safeguard leads to models that frequently produce noisy or artifact-prone images, whereas SDPO-style control yields more stable optimization and systematically better visual quality.

We retain the same learning rate as used in the SFT stage, but adopt a large global batch size and $\beta$ in order to obtain stable preference gradients and avoid drifting far from the supervised baseline. The DPO stage enhances the model by improving its helpfulness, harmlessness, adherence to prompts (including layout and text rendering), and by reducing noticeable artifacts.

\textbf{Stage 3: GRPO.}
After training with DPO, we refine the model using Group Relative Policy Optimization (GRPO) \citep{shao2024deepseekmath, liu2025flow}, conducting on-policy sampling during training and evaluating with a set of reward models. For each prompt, the model generates multiple candidate images as a group, which are then scored by a combination of reward models. Conditioned on a  text prompt $c$, the flow model predicts a group of $G$ individual images $\{x_0^i\}_{i=1}^G$ with their corresponding trajectorys $\{x_T^i, x_{T-1}^i, ..., x_0^i\}_{i=1}^G$ . The advantage of $i$-th image within the group can be formulated as:
\begin{equation}
A_i = \frac{R(x_0^i, c) - mean(\{R(x_0^i, c)\}_{i=1}^G)}{std(\{R(x_0^i, c)\}_{i=1}^G)},
\end{equation}

where $R$ denotes the reward model. Consequently, the training objective of GRPO is:

\begin{equation}
\mathcal{L}_{\text{GRPO}}(\theta) = \mathbb{E}_{c\sim \mathcal{D},\{x_T^i,...,x_0^i\}_{i=1}^G\sim\pi_\theta} f(r,A,\theta,\epsilon,\beta)\\
\end{equation}

where 
\begin{align*}
    f(r,A,\theta,\epsilon,\beta) &= \frac{1}{G} \sum_{i=1}^G \frac{1}{T} \sum_{t=0}^{T-1} \left(\min(r_t^i(\theta) A_i, \text{clip}(r_t^i(\theta), 1-\epsilon, 1+\epsilon) A_i) - \beta D_{KL}(\pi_\theta || \pi_{\text{ref}})\right)\\
    r_t^i(\theta) &= \frac{p_\theta(x_{t-1}^i|x_t^i,c)}{p_{\theta_{\text{old}}}(x_{t-1}^i|x_t^i,c)}.
\end{align*}
To accelerate training with minimizing the impact on performance, we sample each candidate image using fewer denoising steps. Furthermore, we introduce coefficients-preserving sampling \citep{wang2025coefficients} during the GRPO stage to further enhance performance. The training window adaptively learns the needs of different denoise stages.
We retain the same learning rate from the DPO stage and run GRPO for approximately 500 steps. Throughout this process, the policy is optimized to maximize the expected reward, while applying a KL penalty to constrain its divergence from the DPO model.

\section{Evaluation}

We evaluate Ovis-Image on the text-to-image task from two perspectives: text rendering capability and general text-to-image generation capability. To evaluate the text rendering capability, we conduct evaluations on LongText-Bench~\citep{geng2025x}and CVTG-2K~\citep{du2025textcrafter}. To evaluate the general text-to-image generation capability, we present a comprehensive evaluation across multiple public benchmarks, including DPG-Bench~\citep{dpg_bench}, GenEval~\citep{geneval} and OneIG-Bench~\citep{chang2025oneig}. Despite its compact parameter size, Ovis-Image consistently outperforms significantly larger open-source baselines (highlighted in gray), proving that it delivers competitive generation quality with superior parameter efficiency. 

\paragraph{CVTG-2K} Table \ref{tab:CVTG-2K} presents the results of the English rendering evaluation on CVTG-2K~\citep{du2025textcrafter}. This benchmark comprises 2,000 prompts, each demanding the rendering of 2 to 5 English text regions on the generated image. It introduces Word Accuracy, NED and CLIPScore to evaluate the precision of the text rendering. As shown in the table, Ovis-Image achieves the highest overall word accuracy across all regions. Additionally, Ovis-Image obtains the highest NED and CLIPScore, further confirming its superior text rendering capability. 

\paragraph{LongText-Bench} Table \ref{tab:main_longtext-bench} presents the evaluation results on LongText-Bench~\citep{geng2025x}, a benchmark designed to examine the model's capability to accurately render long texts in both English and Chinese. As illustrated in the table, Ovis-Image demonstrates superior performance on the Chinese text. Despite its relatively small model parameter, Ovis-Image still excels at generating long English text, achieving performance comparable against closed-source models and models with larger parameter counts. This result highlights the particular strength of Ovis-Image in long text generation capabilities.

\begin{table}[h]
\centering
\caption{Evaluation of text rendering ability on CVTG-2K.}
\renewcommand{\arraystretch}{1.15} % 调整行高以增加可读性
% 定义列格式：左对齐 | 5个居中 | 1个居中 | 1个居中
\resizebox{\textwidth}{!}{
\label{tab:CVTG-2K}
\begin{tabular}{lc|ccccc|c|c}
\toprule
\multirow{2}{*}{\textbf{Model}}& \#Params. & \multicolumn{5}{c|}{\textbf{Word Accuracy}$\uparrow$} & \multirow{2}{*}{\textbf{NED}$\uparrow$} & \multirow{2}{*}{\textbf{CLIPScore}$\uparrow$} \\
\cline{3-7}
& & 2 regions & 3 regions & 4 regions & 5 regions & average & & \\
\midrule
Seedream 3.0~\citep{gao2025seedream}&- & 0.6282 & 0.5962 & 0.6043 & 0.5610 & 0.5924 & 0.8537 & 0.7821 \\
GPT4o~\citep{chatgpt4o}&- & 0.8779 & 0.8659 & 0.8731 & 0.8218 & 0.8569 & 0.9478 & 0.7982 \\
\midrule
SD3.5 Large~\citep{esser2024scaling} & 11B+8B\enspace & 0.7293 & 0.6825 & 0.6574 & 0.5940 & 0.6548 & 0.8470 & 0.7797 \\
\rowcolor{gray!20} RAG-Diffusion~\citep{chen2024region} & 11B+12B & 0.4388 & 0.3316 & 0.2116 & 0.1910 & 0.2648 & 0.4498 & 0.7797 \\
\rowcolor{gray!20} FLUX.1-dev~\citep{flux2024}&11B+12B & 0.6089 & 0.5531 & 0.4661 & 0.4316 & 0.4965 & 0.6879 & 0.7401 \\
\rowcolor{gray!20} TextCrafter~\citep{du2025textcrafter}&11B+12B & 0.7628 & 0.7628 & 0.7406 & 0.6977 & 0.7370 & 0.8679 & 0.7868 \\
\rowcolor{gray!20} Qwen-Image~\citep{qwen_image}&\enspace 7B+20B & 0.8370 & 0.8364 & 0.8313 & 0.8158 & 0.8288 & 0.9116 & 0.8017 \\
\midrule
Ovis-Image&2B+7B & \textbf{0.9248} & \textbf{0.9239} & \textbf{0.9180} & \textbf{0.9166} & \textbf{0.9200} & \textbf{0.9695} & \textbf{0.8368} \\
\bottomrule
\end{tabular}
}
\end{table}

\begin{table}[h]
  \caption{Evaluation of text rendering ability on LongText-Bench.}
  \label{tab:main_longtext-bench}
  \centering
  \begin{tabular}{lc|c|c}
    \toprule
    \textbf{Model}& \#Params. & LongText-Bench-EN & LongText-Bench-ZN \\
    \midrule
    Kolors 2.0~\citep{kuaishou2025kolors}&- & 0.258 & 0.329 \\
    GPT4o~\citep{chatgpt4o}&-  & \textbf{0.956} & 0.619 \\
    Seedream 3.0~\citep{gao2025seedream}&- & 0.896 & 0.878 \\
    \midrule
    OmniGen2~\citep{wu2025omnigen2}&3B+4B & 0.561 & 0.059 \\
    Janus-Pro~\citep{chen2025janus}&7B & 0.019 & 0.006 \\
    BLIP3-o~\citep{chen2025blip3}&7B+1B & 0.021 & 0.018 \\
    \rowcolor{gray!20} FLUX.1-dev~\citep{flux2024}&11B+12B & 0.607 & 0.005 \\
    \rowcolor{gray!20} BAGEL~\citep{deng2025emerging}&7B+7B & 0.373 & 0.310 \\
    \rowcolor{gray!20} HiDream-I1-Full~\citep{cai2025hidream}&11B+17B & 0.543 & 0.024 \\
    \rowcolor{gray!20} Qwen-Image~\citep{qwen_image}&\enspace7B+20B & 0.943 & 0.946 \\
    \midrule
    Ovis-Image&2B+7B & 0.922 & \textbf{0.964} \\
    \bottomrule
  \end{tabular}
\end{table}

% Despite having fewer total parameters than some competing approaches, Ovis-Image consistently produces more faithful and semantically coherent results, suggesting that its architecture and training strategy effectively capture and generate complex textual content in visual form.

% \textbf{General Text-to-Image Generation.} 

\paragraph{DPG-Bench} Table \ref{tab:main_dpg-bench} reports the results on DPG-Bench~\citep{dpg_bench}, a benchmark of 1,000 dense prompts intended to evaluate the alignment of text-to-image generation in various dimensions, allowing a detailed inspection of prompt adherence from multiple perspectives. Overall, Ovis-Image delivers robust performance compared to both close-source models and open-source models with larger parameter counts.

\paragraph{GenEval} Table \ref{tab:main_geneval} summarizes the performance on GenEval~\citep{geneval} benchmark, which emphasizes object-centric text-to-image generation by employing compositional prompts with a wide range of object attributes. These results exhibit the competitive controllable generation capabilities of Ovis-Image.

\paragraph{OneIG-Bench} Table \ref{tab:main_oneigen-bench} and Table \ref{tab:main_oneigzn-bench} show the performance comparison on OneIG-Bench~\citep{chang2025oneig}, a comprehensive benchmark developed for detailed evaluation of T2I models across multiple dimensions. As shown in the table, Ovis-Image demonstrates exceptional bilingual performance, particularly distinguished by its performance in the text dimensions.

\begin{table}[h]
  \caption{Evaluation of text-to-image generation ability on DPG-Bench.}
  \label{tab:main_dpg-bench}
  \centering
  \resizebox{\textwidth}{!}{
  \begin{tabular}{lc|ccccc|c}
    \toprule
    \textbf{Model}& \#Params. & Global & Entity & Attribute & Relation & Other & Overall \\
    \midrule
    Seedream 3.0~\citep{gao2025seedream}&- & \textbf{94.31} & \textbf{92.65} & 91.36 & 92.78 & 88.24 & 88.27 \\
    GPT4o~\citep{chatgpt4o}&- & 88.89 & 88.94 & 89.84 & 92.63 & 90.96 & 85.15 \\
    \midrule
    Ovis-U1~\citep{ovis_u1}&2B+1B & 82.37 & 90.08 & 88.68 & 93.35 & 85.20 & 83.72 \\
    OmniGen2~\citep{wu2025omnigen2}&3B+4B & 88.81 & 88.83 & 90.18 & 89.37 & 90.27 & 83.57 \\
    Janus-Pro~\citep{chen2025janus}& 7B & 86.90 & 88.90 & 89.40 & 89.32 & 89.48 & 84.19 \\
    \rowcolor{gray!20} BAGEL~\citep{deng2025emerging}&7B+7B & 88.94 & 90.37 & 91.29 & 90.82 & 88.67 & 85.07 \\
    \rowcolor{gray!20} HiDream-I1-Full~\citep{cai2025hidream}&11B+17B & 76.44 & 90.22 & 89.48 & 93.74 & 91.83 & 85.89 \\
    \rowcolor{gray!20} UniWorld-V1~\citep{lin2025uniworld}& \enspace7B+12B & 83.64 & 88.39 & 88.44 & 89.27 & 87.22 & 81.38 \\
    \rowcolor{gray!20} Qwen-Image~\citep{qwen_image}&\enspace7B+20B & 91.32 & 91.56 & \textbf{92.02} & \textbf{94.31} & \textbf{92.73} & \textbf{88.32} \\
    \midrule
    Ovis-Image&2B+7B & 82.37 & 92.38 & 90.42 & 93.98 & 91.20 & 86.59  \\
    \bottomrule
  \end{tabular}
  }
\end{table}

\begin{table}[h]
  \caption{Evaluation of text-to-image generation ability on GenEval.}
  \label{tab:main_geneval}
  \centering
  \resizebox{\textwidth}{!}{
  \begin{tabular}{lc|cccccc|c}
    \toprule
    \textbf{Model}& \#Params. & Single object & Two object & Counting & Colors & Position & Attribute binding & Overall \\
    \midrule
    Seedream 3.0~\citep{gao2025seedream}&- & 0.99 & 0.96 & \textbf{0.91} & \textbf{0.93} & 0.47 & \textbf{0.80} & 0.84 \\
    GPT4o~\citep{chatgpt4o}&- & 0.99 & 0.92 & 0.85 & 0.92 & 0.75 & 0.61 & 0.84 \\
    \midrule
    Ovis-U1~\citep{ovis_u1}&2B+1B & 0.98 & \textbf{0.98} & 0.90 & 0.92 & \textbf{0.79} & 0.75 & \textbf{0.89} \\
    OmniGen2~\citep{wu2025omnigen2}&3B+4B & \textbf{1.00} & 0.95 & 0.64 & 0.88 & 0.55 & 0.76 & 0.80 \\
    Janus-Pro~\citep{chen2025janus} &7B & 0.99 & 0.89 & 0.59 & 0.90 & \textbf{0.79} & 0.66 & 0.80 \\
    \rowcolor{gray!20} BAGEL~\citep{deng2025emerging}&7B+7B & 0.99 & 0.94 & 0.81 & 0.88 & 0.64 & 0.63 & 0.82 \\
    \rowcolor{gray!20} HiDream-I1-Full~\citep{cai2025hidream}&11B+17B & 1.00 & \textbf{0.98} & 0.79 & 0.91 & 0.60 & 0.72 & 0.83 \\
    \rowcolor{gray!20} UniWorld-V1~\citep{lin2025uniworld}&\enspace7B+12B & 0.99 & 0.93 & 0.79 & 0.89 & 0.49 & 0.70 & 0.80 \\
    \rowcolor{gray!20} Qwen-Image~\citep{qwen_image}&\enspace7B+20B & 0.99 & 0.92 & 0.89 & 0.88 & 0.76 & 0.77 & 0.87 \\
    \midrule
    Ovis-Image &2B+7B& \textbf{1.00} & 0.97 & 0.76 & 0.86 & 0.67 & \textbf{0.80} & 0.84 \\
    \bottomrule
  \end{tabular}
  }
\end{table}

\begin{table}[h]
  \caption{Evaluation of text-to-image generation ability on OneIG-EN.}
  \label{tab:main_oneigen-bench}
  \centering
  \resizebox{\textwidth}{!}{
  \begin{tabular}{lc|ccccc|c}
    \toprule
    \textbf{Model} & \#Params. & Alignment & Text & Reasoning & Style & Diversity & Overall \\
    \midrule
    Kolors 2.0~\citep{kuaishou2025kolors}& - & 0.820 & 0.427 & 0.262 & 0.360 & 0.300 & 0.434 \\
    Imagen4~\citep{Imagen}& - & 0.857 & 0.805 & 0.338 & 0.377 & 0.199 & 0.515 \\
    Seedream 3.0~\citep{gao2025seedream}& - & 0.818 & 0.865 & 0.275 & 0.413 & 0.277 & 0.530 \\
    GPT4o~\citep{chatgpt4o}& - & 0.851 & 0.857 & \textbf{0.345} & \textbf{0.462} & 0.151 & 0.533 \\
    \midrule
    Ovis-U1~\citep{ovis_u1}& 2B+1B & 0.816 & 0.034 & 0.226 & 0.443 & 0.191 & 0.342 \\
    CogView4~\citep{Cogview4}& 6B & 0.786 & 0.641 & 0.246 & 0.353 & 0.205 & 0.446 \\
    Janus-Pro~\citep{chen2025janus} & 7B & 0.553 & 0.001 & 0.139 & 0.276 & \textbf{0.365} & 0.267 \\
    OmniGen2~\citep{wu2025omnigen2}& 3B+4B & 0.804 & 0.680 & 0.271 & 0.377 & 0.242 & 0.475 \\
    BLIP3-o~\citep{chen2025blip3}& 7B+1B & 0.711 & 0.013 & 0.223 & 0.361 & 0.229 & 0.307 \\
    \rowcolor{gray!20} FLUX.1-dev~\citep{flux2024}& 11B+12B & 0.786 & 0.523 & 0.253 & 0.368 & 0.238 & 0.434 \\
    \rowcolor{gray!20} BAGEL~\citep{deng2025emerging}& 7B+7B & 0.769 & 0.244 & 0.173 & 0.367 & 0.251 & 0.361 \\
    \rowcolor{gray!20} BAGEL+CoT~\citep{deng2025emerging}& 7B+7B & 0.793 & 0.020 & 0.206 & 0.390 & 0.209 & 0.324 \\
    \rowcolor{gray!20} HiDream-I1-Full~\citep{cai2025hidream}& 11B+17B & 0.829 & 0.707 & 0.317 & 0.347 & 0.186 & 0.477 \\
    \rowcolor{gray!20} HunyuanImage-2.1~\citep{HunyuanImage-2.1}& \enspace7B+17B   & 0.835 & 0.816 & 0.299 & 0.355 & 0.127 & 0.486 \\
    \rowcolor{gray!20} Qwen-Image~\citep{qwen_image}&\enspace7B+20B & \textbf{0.882} & 0.891 & 0.306 & 0.418 & 0.197 & \textbf{0.539} \\
    \midrule
    Ovis-Image& 2B+7B & 0.858 & \textbf{0.914} & 0.308 & 0.386 & 0.186 & 0.530 \\
    \bottomrule
  \end{tabular}
  }
\end{table}

\begin{table}
  \caption{Evaluation of text-to-image generation ability on OneIG-ZN.}
  \label{tab:main_oneigzn-bench}
  \centering
  \resizebox{\textwidth}{!}{
  \begin{tabular}{lc|ccccc|c}
    \toprule
    \textbf{Model} & \#Params. & Alignment & Text & Reasoning & Style & Diversity & Overall \\
    \midrule
    Kolors 2.0~\citep{kuaishou2025kolors} & - & 0.738 & 0.502 & 0.226 & 0.331 & 0.333 & 0.426 \\
    Seedream 3.0 \citep{gao2025seedream} & - & 0.793 & 0.928 & 0.281 & 0.397 & 0.243 & 0.528 \\
    GPT4o~\citep{chatgpt4o} & - & 0.812 & 0.650 & \textbf{0.300} & \textbf{0.449} & 0.159 & 0.474 \\
    \midrule
    CogView4~\citep{Cogview4} & 6B & 0.700 & 0.193 & 0.236 & 0.348 & 0.214 & 0.338 \\
    Janus-Pro~\citep{chen2025janus} & 7B & 0.324 & 0.148 & 0.104 & 0.264 & \textbf{0.358} & 0.240 \\
    BLIP3-o~\citep{chen2025blip3} & 7B+1B & 0.608 & 0.092 & 0.213 & 0.369 & 0.233 & 0.303 \\
    \rowcolor{gray!20} BAGEL~\citep{deng2025emerging} & 7B+7B & 0.672 & 0.365 & 0.186 & 0.357 & 0.268 & 0.370 \\
    \rowcolor{gray!20} BAGEL+CoT~\citep{deng2025emerging} & 7B+7B & 0.719 & 0.127 & 0.219 & 0.385 & 0.197 & 0.329 \\
    \rowcolor{gray!20} HiDream-I1-Full~\citep{cai2025hidream} & 11B+17B & 0.620 & 0.205 & 0.256 & 0.304 & 0.300 & 0.337 \\
    \rowcolor{gray!20} HunyuanImage-2.1~\citep{HunyuanImage-2.1} & \enspace7B+17B & 0.775 & 0.896 & 0.271 & 0.348 & 0.114 & 0.481 \\
    \rowcolor{gray!20} Qwen-Image~\citep{qwen_image} & \enspace7B+20B & \textbf{0.825} & \textbf{0.963} & 0.267 & 0.405 & 0.279 & \textbf{0.548} \\
    \midrule
    Ovis-Image & 2B+7B & 0.805 & 0.961 & 0.273 & 0.368 & 0.198 & 0.521 \\
    \bottomrule
  \end{tabular}
  }
\end{table}

\paragraph{Computational Overhead} Table \ref{tab:text_to_image} presents a comparative analysis of computational overhead, focusing on inference time and GPU memory utilization. Ovis-Image exhibits a superior trade-off between resource efficiency and model performance. Most notably, Ovis-Image maintains a significantly lower memory footprint compared to larger baselines. Furthermore, in terms of temporal efficiency, Ovis-Image achieves a substantial speedup,thereby offering a more practical solution for resource-constrained environments.

\begin{table}[h!]
\centering
\caption{Comparison of model inference time and GPU memory usage (1024×1024 images, 50-step sampling, BF16 inference)}
\label{tab:text_to_image}
\begin{tabular}{lc|c|cc|cc}
\toprule
\multirow{2}{*}{Model} & \multirow{2}{*}{\#Params.} & \multirow{2}{*}{Accelerate} &\multicolumn{2}{c}{A100} & \multicolumn{2}{c}{H100} \\
\cmidrule(lr){4-5} \cmidrule(lr){6-7}
& & & Memory (MB) & Time (s) & Memory (MB) & Time (s) \\
\midrule
Flux.1-dev & 11B+12B & guidance dis. & 34637   & 23.51 & 34661    & 11.03 \\
Qwen-Image & \enspace 7B+20B & - & 59329   & 45.16 & 59354    & 20.27 \\
Ovis-U1 &  2B+1B    & - & 10528   & 8.41  & 11937 & 4.29  \\
Ovis-Image& 2B+7B  & - & 24959   & 30.56 & 24276    & 13.74 \\
\bottomrule
\end{tabular}
\end{table}

\section{Conclusion}
\label{sec:conclusion}

We presented Ovis-Image, a 7B text-to-image model designed to reconcile strong in-image text rendering with practical deployment cost. By pairing a diffusion-based visual decoder with the Ovis 2.5 multimodal backbone and training it through a text-centric pipeline, Ovis-Image attains text rendering quality comparable to much larger open models and approaches leading closed-source systems, while preserving solid general-purpose generation and fitting on a single high-end GPU. Beyond the empirical gains, Ovis-Image illustrates a more general design principle: frontier-like text-aware generation can emerge from moderate-scale models when architectural choices, data curation, and alignment objectives are explicitly organized around the demands of in-image text, rather than treated as a byproduct of generic image synthesis.

\section{Contributors}

Guo-Hua Wang\footnote{Correspondence to Guo-Hua Wang $<$\texttt{wangguohua@alibaba-inc.com}$>$}, Liangfu Cao, Tianyu Cui, Minghao Fu\footnote{Work done during the internship at Alibaba Group\label{intern}}, Xiaohao Chen, Pengxin Zhan,  Jianshan Zhao, Lan Li\footref{intern}, Bowen Fu\footref{intern}, Jiaqi Liu, Qing-Guo Chen

% Below, we provide an alphabetical list of contributors to Ovis-Image:

% \begin{itemize}
%     \item \textbf{Project Leader:} Qing-Guo Chen
%     \item \textbf{Data:} Jianshan Zhao, Jiaqi Liu, Liangfu Cao, Xiaohao Chen, 
%     \item \textbf{Infra: } Guo-Hua Wang, Lan Li
%     \item \textbf{Model Architecture \& Pretraining: } Guo-Hua Wang\footnote{Correspondence to Guo-Hua Wang $<$\texttt{wangguohua@alibaba-inc.com}$>$}
%     \item \textbf{Post Training: } Guo-Hua Wang, Minghao Fu, Tianyu Cui
%     \item \textbf{Evaluation: } Bowen Fu, Liangfu Cao, Pengxin Zhan
% \end{itemize}

% \clearpage

\bibliographystyle{colm2024_conference}
\bibliography{colm2024_conference}

\end{document}